\documentclass[letterpaper]{article} 
\usepackage{aaai24}  
\usepackage{times}  
\usepackage{helvet}  
\usepackage{courier}  
\usepackage[hyphens]{url}  
\usepackage{graphicx} 
\usepackage{natbib}  
\usepackage{caption} 
\usepackage{algorithm}
\usepackage{algorithmic}

\usepackage{newfloat}
\usepackage{listings}
\DeclareCaptionStyle{ruled}{labelfont=normalfont,labelsep=colon,strut=off} 
\floatstyle{ruled}
\newfloat{listing}{tb}{lst}{}
\floatname{listing}{Listing}
\title{GameEval: Evaluating LLMs on Conversational Games}

\author{\small Dan Qiao $^{1}$\thanks{\scriptsize Both authors contributed equally to this research.} \quad Chenfei Wu$^{2}$\samethanks[1] \quad Yaobo Liang$^{2}$ \quad  Juntao Li$^{1}$ \quad \textbf{Nan Duan}$^{2}$\thanks{\scriptsize Corresponding author.} \\
{$^{1}$Soochow University\quad $^{2}$Microsoft Research Asia} \\
{\tt\small \{dqiaojordan@stu.,ljt@\}suda.edu.cn, \{chewu,yaobo.liang,nanduan\}@microsoft.com}
}

\newcommand*\samethanks[1][\value{footnote}]{\footnotemark[#1]}
\usepackage{bibentry}
\usepackage{graphicx}
\usepackage{subfigure}
\usepackage{amsmath}
\usepackage{multirow}
\usepackage{makecell}
\usepackage{booktabs}
\usepackage{bbding}

\begin{document}

\maketitle
\begin{abstract}

The rapid advancements in large language models (LLMs) have presented challenges in evaluating those models. 
Existing evaluation methods are either reference-based or preference based, which inevitably need human intervention or introduce test bias caused by evaluator models.
In this paper, we propose GameEval, a novel approach to evaluating LLMs through goal-driven conversational games, overcoming the limitations of previous methods. 
GameEval treats LLMs as game players and assigns them distinct roles with specific goals achieved by launching conversations of various forms, including discussion, question answering, and voting.
We design three unique games with cooperative or adversarial objectives, accompanied by corresponding evaluation metrics, to show how this new paradigm comprehensively evaluates model performance.
Through extensive experiments, we show that GameEval can effectively differentiate the capabilities of various LLMs, providing a comprehensive assessment of their integrated abilities to solve complex problems.
Our public anonymous code is available at https://github.com/GameEval/GameEval.
\end{abstract}

\section{Introduction}
Starting from the pre-trained models like BERT \cite{bert}, GPT \cite{gpt} to the recent powerful Large Language Models (LLMs) such as LLaMA \cite{llama}, Vicuna \cite{vicuna} and GPT-4 \cite{gpt4}, we have witnessed a great evolution of the capabilities of LLMs, which can now perform zero-shot generation, reason for complex tasks, and follow human instructions. 
With the newly emerged capabilities and the universal task generalization, the evaluation of LLMs has shifted dramatically from the previous widely used benchmarks like GLUE \cite{GLUE}, SuperGLUE\cite{SuperGLUE}, MMLU \cite{MMLU} to recent general and open-ended evaluation, e.g., AGIEval \cite{agieval}, Chatbot-Arena \cite{MT-Bench}.

\begin{figure}[t]
\centering

    \centering
    \includegraphics[width=0.47\textwidth]{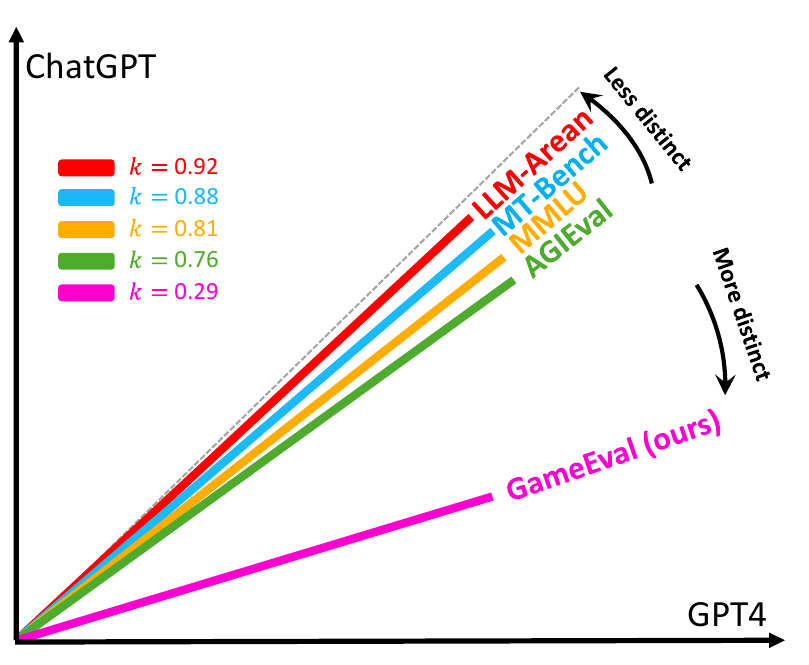}

\centering

\caption{Comparison between our proposed GameEval and the widely used benchmarks. The slope represents the ratio between the performance score of ChatGPT and GPT-4. By playing goal-driven conversational games, GameEval provides more distinguishable results in ChatGPT vs GPT-4.}
\label{fig:cmp}
\vspace{-6mm}
\end{figure}

Existing evaluation methods primarily fall into two categories: reference-based or preference-based. Reference-based methods require a ground-truth answer, either in the form of labels or referenced text \cite{GLUE, SuperGLUE, agnews, agieval, bigbench, MMLU}. 
However, acquiring high-quality annotations can be expensive and time-consuming.
Furthermore, for complex scenarios or open-ended tasks, it is unreasonable to only compare the output with the standard answer, as multiple valid solutions may exist.
In contrast, preference-based methods employ humans or models as evaluators.
Although these methods can assess LLMs' real-world performance, they either require substantial human resources or introduce preference bias from the evaluator models \cite{alpaca-eval}.
More importantly, current evaluation methods assess multiple tasks individually, with the final score derived from an average of these separate evaluations. 
Since each task corresponds to only specific capabilities of the model, these evaluations can not effectively assess the model's ability to utilize all the capabilities simultaneously.
Consequently, this makes it more challenging to distinguish the real-world performance of LLMs
As illustrated in Figure~\ref{fig:cmp}, when comparing the performance of ChatGPT and GPT-4, which has been shown to exhibit significant differences in GPT-4 technical report ~\cite{gpt4}, the obtained results are still relatively close (with a slope close to 1). 
This evidence suggests that more comprehensive and in-depth evaluation methods are necessary to accurately assess and compare the real-world performance of these models on complex tasks.

Building upon the observations mentioned above, we present GameEval, a pioneering approach designed to assess the integrated capabilities of LLMs by engaging them in goal-driven conversational games. 
In GameEval, LLMs are treated as players and assigned distinct roles, each with a specific goal to achieve to win the game. The dialogue process encompasses various forms, including discussion, question answering, and voting. 
The outcomes of the games effectively reflect whether the model's outputs contribute to achieving the assigned goal, thus serving as a direct quantification of the model's capabilities.
To accomplish the goals of their assigned roles, models must demonstrate high-level competencies in multiple aspects, including but not limited to cooperation, multi-hop reasoning, summarizing, instruction-following, and long-term planning. 
Our proposed GameEval is neither based on reference nor preference, eliminating the need for human intervention and reducing test bias caused by evaluator models. 
Unlike existing evaluation methods that average scores from multiple individual tasks, GameEval requires models to apply these capabilities simultaneously in each conversation round to achieve the game's long-term goal. 
This offers a more comprehensive assessment of LLMs' integrated capabilities to solve complex problems. Consequently, GameEval is capable of significantly differentiating LLMs. For instance, Figure \ref{fig:cmp} demonstrates the distinction between ChatGPT and GPT-4, with a slope of 0.29 between them.
Our main contributions can be summarized as follows:

\begin{itemize}
    
    \item We propose GameEval, the first paper to evaluate LLMs by playing goal-driven conversational games, which can provide a comprehensive assessment of LLMs' integrated capabilities while eliminating testing bias and ground-truth label reliance.
    
    \item We designed three games with adversarial or cooperative objectives, along with corresponding evaluation metrics. These games can be utilized to assess the performance of various generative large language models.
    
    \item We conducted extensive experiments, and we observed significant differences among various models, effectively highlighting the distinction in their capabilities.
    
\end{itemize}

\section{Related Work}

\subsection{Reference-based Evaluation}
Reference-based evaluation methods aim to evaluate the performance of an LLM by comparing the model's output results with the pre-determined reference \cite{GLUE, bigbench, SuperGLUE, API-Bank, math}. 
For NLU tasks, the reference is a corresponding correct answer, and the model's outputs are expected to be the same as the given reference.
For generation tasks, the performance of the model can be evaluated by calculating the similarity to the reference text. 
BLUE \citet{BLEU}, ROUGE \citet{rouge}, and BERTscore \citet{bert} are commonly used for measuring the similarity between the output and test reference text. 

With the development of large language models, single tasks can no longer comprehensively evaluate the capabilities of the models, and it's natural to test the model on multiple datasets of diverse domains in various tasks \cite{openllm,c-eval,promptbench,m3ke}. 
For example, GLUE \cite{GLUE} and SuperGLUE \cite{SuperGLUE} are the representative multi-task evaluation benchmarks aiming to emulate authentic language processing scenarios,  covering a wide range of tasks such as text classification, machine translation, reading comprehension, and dialogue generation. 
\citet{MMLU} developed the MMLU benchmark to evaluate the acquisition and application of knowledge obtained during pre-training by requiring the model to answer multiple-choice questions from 57 diverse tasks. These tasks encompass various domains such as elementary mathematics, US history, computer science, law, and more.
AGIEval \cite{agieval} assesses foundation models in the context of human-centric standardized exams like college entrance exams and law school tests.
In addition to the benchmarks designed for the general domain, works like \citet{math} and \citet{toolbench} evaluate the capabilities of solving math problems and using tools.
Although reference-based evaluation offers an objective and reliable assessment, making it widely utilized in model evaluation, it still presents certain limitations.
For one thing, acquiring high-quality references entails significant costs.
For another, in complex scenarios like open-ended generation where multiple correct answers exist, merely comparing the model's output with the reference answer has limitations.
Moreover, LLMs have read massive corpus during pre-training. If the test set overlaps with the pre-trained corpus, it will also introduce test time bias.

\subsection{Preference-based Evaluation} 
To better evaluate the performance of LLMs under complex scenarios, especially for real-world application, many works employ preferences provided by humans or other powerful models to compare the performance of different LLMs.

Human-based evaluation takes the human's preference to judge the performance of LLMs \cite{human1,chatbot,MT-Bench,human2,human3}.
\citet{human1} conducted human evaluations in summarization and disinformation scenarios on six distinct models.
Chatbot Arena \cite{MT-Bench} is one of the most famous benchmarks which introduce anonymous, randomized battles between pairwise models. A user can chat with two anonymous models side-by-side and vote for the better one he thinks. It adopts the Elo rating system to give the final score of participating models.
Considering the high cost of manual evaluation, recent works use a powerful LLM as an alternative to scoring the output of models under evaluation \cite{llm-eval}.
MT-Bench \cite{MT-Bench} offers an extensive collection of meticulously crafted questions aimed at evaluating the proficiency of models in managing multi-turn dialogues. They employ GPT-4 as a surrogate for humans to select the better output generated by participating models. 
Alpaca-Eval \cite{alpaca-eval} focuses on models' instruction-following ability across various natural language processing tasks. It evaluates models by measuring the proportion that a powerful LLM prefers its outputs over the outputs from a reference model.
Vicuna-Eval \cite{vicuna} uses GPT-4 to create a series of questions and asks GPT-4 to rate the quality of the models' output based on helpfulness, relevance, accuracy, and details.
However, preference bias exists in model-based evaluation.
Automatic evaluators (e.g., GPT-4) prefer models that generate longer outputs and/or that were fine-tuned on the model underlying the evaluator \cite{alpaca-eval}.

In contrast with the aforementioned methods, the proposed GameEval eliminates the need for both collected reference and preference supplied by humans or alternative models, introducing a new paradigm for LLM evaluation.
In GameEval, models are required to use multiple abilities simultaneously in each round of conversation to achieve the final goal, providing a more comprehensive assessment of LLMs' integrated capabilities to solve complex problems.

\section{GameEval}

\begin{figure*}[t]
\centering
\subfigure[ask-guess]{
    \begin{minipage}[t]{0.3\linewidth}
        \centering
        \includegraphics[width=\textwidth]{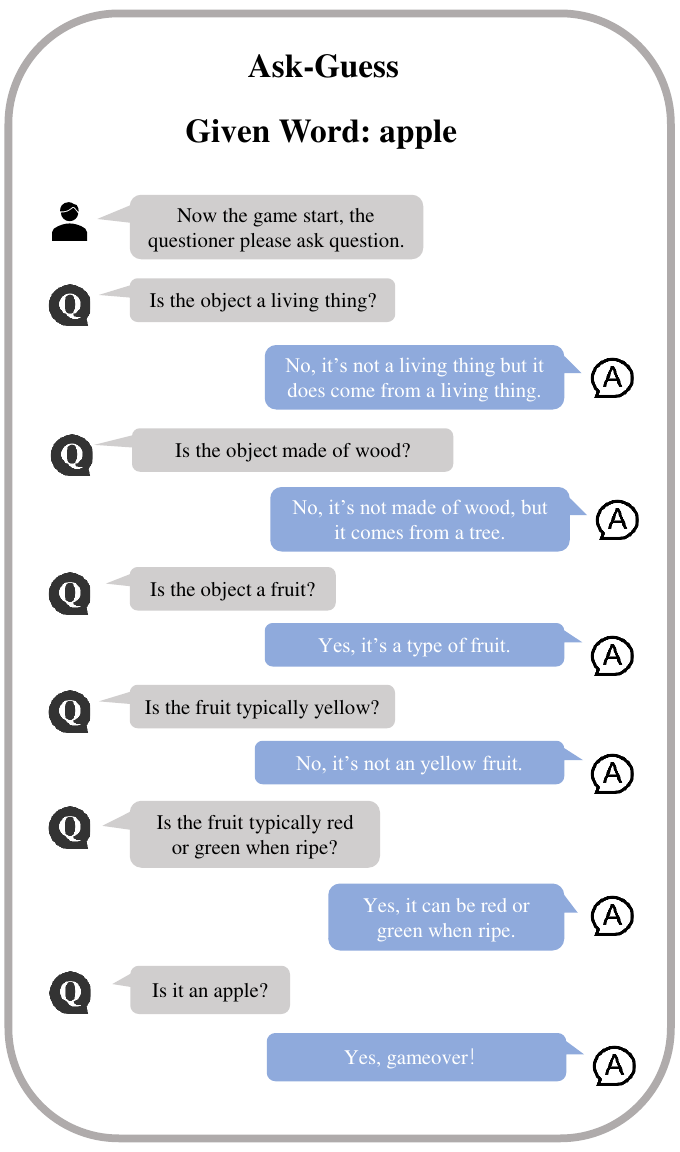}
    \end{minipage}
    \label{mainfig:1}
}
\subfigure[spyfall]{
    \begin{minipage}[t]{0.3\linewidth}
        \centering
        \includegraphics[width=\textwidth]{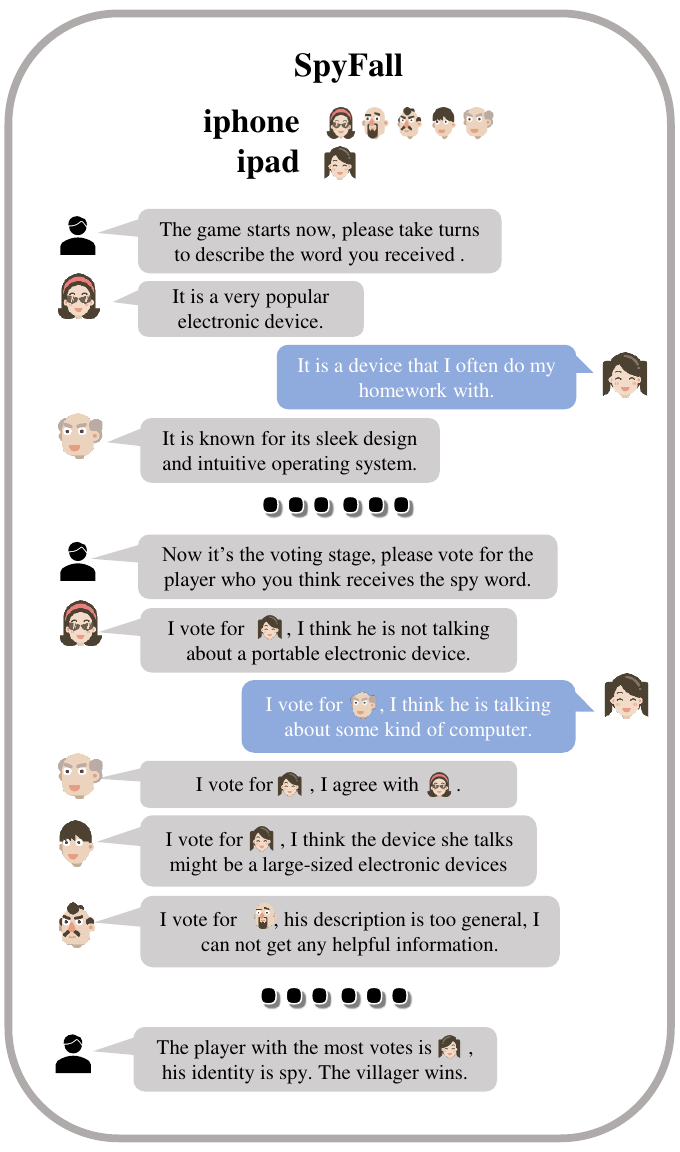}
    \end{minipage}
    \label{mainfig:2}
} 
\subfigure[tofukingdom]{
    \begin{minipage}[t]{0.3\linewidth}
        \centering
        \includegraphics[width=\textwidth]{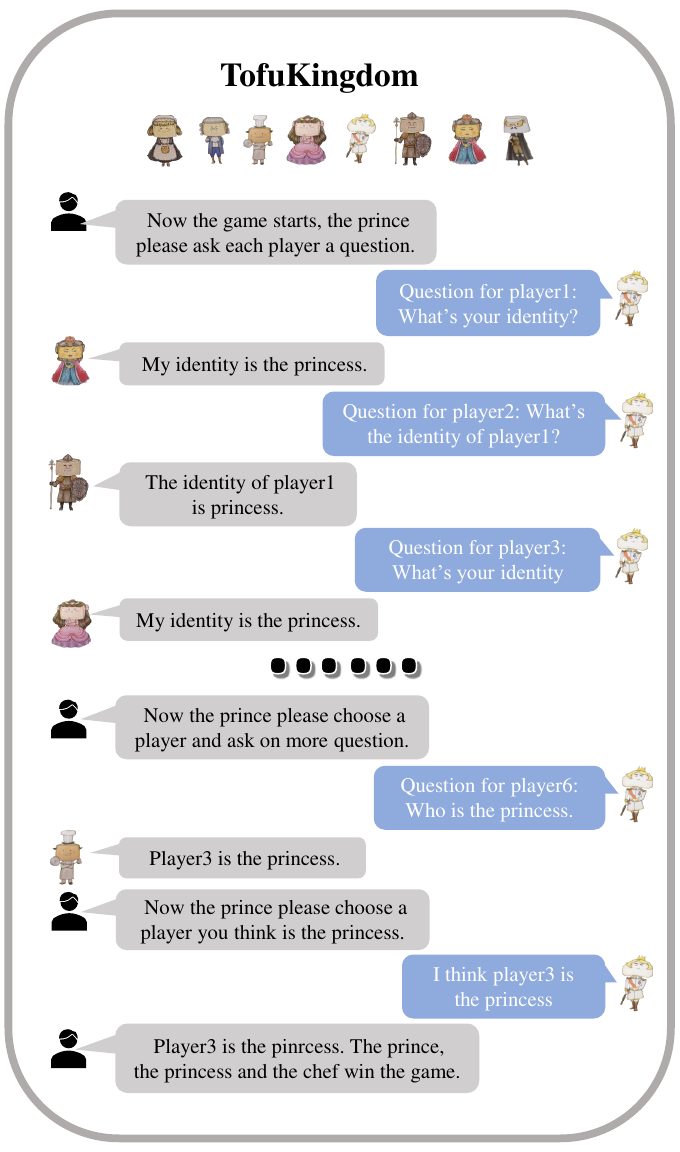}
    \end{minipage}
    \label{mainfig:3}
} 

\centering
\caption{(a) An example of the game Ask-Guess, where the given word is ``apple.'' (b) An example of the game SpyFall, where the common word is ``iphone,'' and the spy word is ``ipad.'' (c) An example of the game TofuKingdom.}
\label{fig:ds-distribution}
\end{figure*}
In this section, we present our game-based evaluation method, where the models' performance is assessed according to the outcome of the game.
We designed three goal-driven conversational games to show how this new paradigm comprehensively evaluates model performance.
The gameplay of these three games ranges from simple to complex, and the number of roles in the games also varies from few to many, each with a cooperative or adversarial long-term goal.
In the following sections, we will provide an introduction to each of these games individually.

\subsection{Ask-Guess}

\subsubsection{Game Introduction}\label{subsec:ask-guess}
Ask-Guess is a cooperative game involving a questioner and an answerer. 
At the beginning of the game, the answerer receives a word unknown to the questioner. 
In each round, the questioner may ask the answerer one question, and the answerer has to answer faithfully.
The provided word or phrase must not be included in the answerer's reply. 
Both participants should collaborate to minimize the number of Q\&A rounds needed for the questioner to deduce the given word or phrase accurately. 
The questioner should ask targeted questions to progressively narrow down the potential scope of the given word based on the answerer's responses. 
The answerer must assess whether the questioner has successfully identified the word and respond with 'Gameover' to conclude the game.

\subsubsection{Evaluation}
For the game Ask-Guess, we selected the 100 labels from the Cifar-100 dataset as the words to be guessed.
The different outcomes of the game will serve as important indicators for evaluating the model's capabilities:
\begin{itemize}
    \item \textbf{Successful Trial (ST)}: This indicates that the model correctly guesses the word within the limited rounds of Q\&A without violating the game rules.
    \item \textbf{Ending Error (EE)}: For a given word, the answerer may output "gameover" even if the questioner hasn't correctly guessed the result, then the game is wrongly terminated.
    \item \textbf{Round Limit Error (RLE)}: We set a round limit for Q\&A. This error occurs when the model has not correctly guessed the word within 30 rounds.
    \item \textbf{Answer Mentioned Error (AME)}: For a given word, if the answerer directly mentions the word in his response, which violates the rule, the game will be terminated.
    \item \textbf{Chat Error (CE)}: This indicates that an error occurred when generating the response, mainly due to API request errors in the experiment.
\end{itemize}
We recorded the occurrence of the above situations in 100 trials for each word and calculated the average for all words. 
Additionally, for successful trials, we record the average number of rounds to guess the answer. We also provide two distinct game settings. 
In one setting, the answerer was allowed to provide a brief description of the word at the beginning of the game. In the other setting, no prior description is provided by the answerer. 
Intuitively, the questioner may require more rounds of Q\&A to guess the answer when no description is provided. 
For the models under evaluation, a higher correct ratio signifies better performance in the game. When the correct ratios are similar, fewer rounds of Q\&A indicate stronger reasoning abilities of the model. 
For other indicators, smaller values represent better robustness.

\subsection{SpyFall}\
\subsubsection{Game Introduction}
This game has six players, including one spy and five villagers.
At the beginning of the game, everyone will receive a word.
The spy will receive the spy word, and others will receive the common word.
Spy word is different but relevant to the common word. For example, the spy word can be "lion," and the common word is "tiger."
There are two stages in each round of the game.
In the first stage, everyone needs to describe the word he got but cannot say the given word directly.
The challenging part of the game is that since each player does not know others' words, so he does not know whether he got a spy word or a common word.
In the second stage, everyone should vote for a player he thinks is the spy according to the descriptions in the first stage and state why he thinks this player is a spy.
The player with the most votes will be eliminated from the game.
If the spy is eliminated, the villagers win.
If less than three players are left, and the spy still lives, then the spy wins the game.
During the game, it is very important to extrapolate the words according to others' descriptions.
Since the spy word and the common word are always relevant, players need to discern subtle distinctions between descriptions.
What's more, some deceptive strategy is required when a player realizes he is the spy.

\subsubsection{Evaluation Metrics}
We adopt apt adversarial game modes to test the performance between pairs of models.
For an \textbf{ordered model pair} $(M_a,M_b)$, we assign $M_a$ the role of the spy who receives the spy $M_b$ the role of other villager players.
For each model pair under evaluation, we record the following indicators to represent the model's capabilities.
\begin{itemize}
    \item \textbf{Spy Winning rate}: Suppose we give $M_a$ the spy word and use the model pair $(M_a,M_b)$ to play the game for $n$ times, $M_a$ win $s_a$ times, than record the winning rate of $M_a$ against $M_b$ as $w_{ab}=s_a / n$
    \item \textbf{Spy Living Round}: Besides winning rate, we also record the round that model $M_a$ lives against $M_b$ when playing the role of the spy as $l_{ab}$.
\end{itemize}
We select 11 pairs of related words as spy words and common words and compute the average score of the mentioned metrics.
For the pair of models $M_a,M_b$, $w_{ab} > w_{ba}$ indicate that model $a$ performed better in this game than model $b$.

\subsection{TofuKingdom}
\subsubsection{Game Introduction}
This game is a role-playing text reasoning game.
It has eight roles, including Prince, Princess, Queen, Minister, Chef, Guard, Maid, and Spy.
The players, except the Prince, know the real identity of the rest of the players.
The Prince needs to guess which player is the Princess by asking one question to each player.
During the game, the Prince's question can only be chosen from the three questions below: 
1. Who is the Princess;
2. What is your identity;
3. What is the identity of \{player\_name\}.

There are three different camps in this game.
The Princess and Chef belong to the Prince Camp; they must tell the truth when answering the question.
The Queen, Minister, and Guard belong to the Queen Camp; they must tell a lie when answering the question.
The Spy and the Maid belong to the Spy Camp and can choose to speak the truth or lie.
After asking each player one question, the Prince can still choose one player to ask an extra question.
The question should also be chosen from one of the three questions mentioned above.
Then the Prince has to choose a player who he thinks is the Princess.
If the Prince correctly chooses Princess, the Chef and the Princess win.
If the Prince chooses the Queen, the Queen, Minister, and Guard win.
If the Prince chooses a player whose identity is neither the Princess nor the Queen, the Maid and Spy wins. Players in the same camp should cooperate with each other against players in other camps

\subsubsection{Evaluation Metrics}
Unlike Ask-Guess and SpyFall, TofuKingdom does not need any pre-defined word or phrase.
We only need to assign game roles to each participating model.
Compared to other games, this game has more roles, and the actions of the roles are more complex.
If there are different LLMs in the same winning camp, it is difficult to analyze which LLM actually contributes to the victory from the game results.
So we assign the same LLM plays all the roles in the same camp and assign different LLMs for different camps.
In a single game, the model playing the roles in the winning camp can obtain one point.
After playing multiple games in different factions, the model with the higher score is considered to perform better.

\subsection{Involved Capabilities}
GameEval is distinct from other evaluation methods, as it requires not only the model's common capabilities like instruct-following but also the model's higher-level skills, including cooperative\&adversarial strategies, and even deceptive strategies and long-term planning. 
In this section, we introduce various distinctive capabilities that can be effectively evaluated by conversational games. 
Table \ref{table:1} shows the capabilities of LLMs that can be examined by these games.
\begin{table}[ht]\small
\centering
\renewcommand{\arraystretch}{1.2}
  \renewcommand\tabcolsep{5pt}
  \begin{tabular}{c|ccc}
  
    \toprule
        Capabilities & Ask-Guess & SpyFall  & TofuKingdom   \\
    \hline
         Cooperative Strategy       & \Checkmark    & \Checkmark  & \Checkmark  \\
         Adversarial Strategy        & \XSolidBrush  & \Checkmark    & \Checkmark  \\
         Specific Knowledge & \Checkmark    & \Checkmark    & \XSolidBrush  \\
         Multi-hop Reasoning& \Checkmark    & \Checkmark  & \Checkmark  \\    
         Deceptive Strategy & \XSolidBrush & \Checkmark  & \Checkmark  \\
         Long-term Planning & \Checkmark     & \Checkmark    & \XSolidBrush \\
         Instruct-Following & \Checkmark    & \Checkmark    & \Checkmark  \\
  \bottomrule
\end{tabular}
  \caption{Capabilities that can be examined by these games.}
\label{table:1}
\vspace{-4mm}
\end{table}

\begin{itemize}
    \item \textbf{Cooperative Strategy}: In such games, players have to cooperate with one another to achieve the same goal. As illustrated in Figure \ref{mainfig:3}, Queen and the Guard are in the same camp in TofuKingdom. When the Queen deceives the Prince by claiming to be the Princess, the Guard should also assert that the Queen is the Princess.
    \item \textbf{Adversarial Strategy}: In the game, players encounter conflicts of interest with their opponents. To gain an advantage, they may disrupt the decision-making processes of other players through the use of deceptive statements.
    \item \textbf{Specific Knowledge}: The game may require some specific knowledge. For example, to play the game Ask-Guess, an LLM needs to have a wealth of knowledge related to the given word. In SpyFall, the players need to discern the subtle differences between descriptions to deduce the spy word and the common word. For example, in Ask-Guess, players need to combine multiple past Q\&A sessions to eliminate impossible assumptions and ask helpful questions.
    \item \textbf{Multi-hop Reasoning}: The models are required to exert a strong reasoning ability to accomplish the final goal. Different from widely used simple reasoning tasks, this kind of reasoning is often multi-hop, which is based on not only the game rules but also the statements of multiple players in past conversations.
    \item \textbf{Deceptive Strategy}: When playing certain roles in the game, the model may have to utilize some deceptive strategy to hide his identity or even tell a lie to mislead other players. In TofuKingdom, roles in the Queen Camp or Spy Camp may tell a lie to disturb the Prince.
    \item \textbf{Long-term Planning} Player should make a long-term plan rather than make a casual speech. As shown in figure \ref{fig:case}, in ask-guess, the questioner can quickly achieve their goal by asking well-planned questions.
    \item \textbf{Instruction-Following}: Instruction-following is a fundamental capability for the state-of-the-art LLM. The instructions are diverse in these games. For example, it can be simple and instant instructions from the host like ``Please ask the question.'' or ``Please vote for the spy.'' It can also be long-term instructions like the game rules or goal of the assigned role. Unlike traditional instruction following tests, models have to follow these instructions at the same time, including instant commands and long-term instructions containing the assigned goal.
\end{itemize}

\begin{table}[ht]\small
\centering
\renewcommand{\arraystretch}{1.2}
  \renewcommand\tabcolsep{5pt}
  \begin{tabular}{l|c|ccccc}
    \toprule
     \multirow{2}*{Model} & \multirow{2}*{Round} &\multicolumn{5}{c}{Details} \\
     & & ST & EE & RLE &  AME & CE \\
    \midrule

     \multicolumn{7}{c}{w/ prior description} \\
      \midrule
    TD003 & 4.39  & 82.71 & 9.47   &  1.84 & 5.97 & 0.01 \\
    ChatGPT & 6.01 & 53.39 & 8.13 & 14.63  & 23.21 & 0.64 \\
    GPT4 & 1.57 & 97.69 & 0.80 &  1.01 & 0.47 & 0.03 \\
      \midrule
    
     \multicolumn{7}{c}{w/o prior description} \\
      \midrule
    TD003 & 15.13  & 42.36 & 19.18 & 37.19  & 0.36 & 0.91 \\
    ChatGPT & 13.78  & 40.50 & 3.88 & 49.89  & 4.57 & 1.16 \\
    GPT4 & 4.01 & 92.77 & 2.95 &  0.84 & 2.75 & 0.69 \\

  \bottomrule
\end{tabular}
  \caption{The performance of TD003, ChatGPT and GPT-4 in the game Ask\&Guess}
\label{table:Ask&Guess}
\end{table}

\begin{table}[ht]\small
\centering
\renewcommand{\arraystretch}{1.2}
  \renewcommand\tabcolsep{5pt}
  \begin{tabular}{ll|cc|cc|cc}
  
    \toprule
        & & \multicolumn{6}{c}{Villager} \\
        & & \multicolumn{2}{c}{TD003} & \multicolumn{2}{c}{ChatGPT}  & \multicolumn{2}{c}{GPT4} \\
        \hline
         & & $w$ & $l$ & $w$ & $r$  & $w$ & $l$ \\
    \hline
     \multirow{3}*{Spy} 
        & TD003   & -       & -     & 0.37 & 2.50 & 0.14 & 1.77 \\
        & ChatGPT    & 0.51  & 2.67 & -      & -     & 0.10 & 1.93 \\
        & GPT4    & 0.70  & 2.88 & 0.48 & 2.67 & -      & -     \\
  \bottomrule
\end{tabular}
  \caption{The performance of TD003, ChatGPT and GPT-4 when play the spy against other models in SpyFall. $w$ is the winning rate, and $l$ is the average living round of the spy.}
\label{table:spyfall}
\end{table}

  

\begin{table}[ht]\small
\centering
\renewcommand{\arraystretch}{1.2}
  \renewcommand\tabcolsep{5pt}
  \begin{tabular}{ccc|ccc}
  
    \toprule
        \multicolumn{3}{c|}{Camps} & \multicolumn{3}{c}{Points} \\
        Prince & Spy & Queen & ChatGPT & GPT-4  & TD003 \\
    \hline
         TD003 & GPT-4 & ChatGPT   & 7  & 9  &  4  \\
         TD003 & ChatGPT & GPT-4   & 5  & 11 & 4  \\
         ChatGPT & GPT-4 & TD003   & 8  & 7  & 5  \\
         ChatGPT & TD003 & GPT-4   & 5  & 9  & 6  \\
         GPT-4 & TD003 & ChatGPT   & 6  & 7  & 7  \\
         GPT-4 & ChatGPT & TD003   & 8  & 8  & 4 \\
    \hline
    \multicolumn{3}{c|}{Total Points} & 39 & 51 & 30 \\
  \bottomrule
\end{tabular}
  \caption{The performance of TD003, ChatGPT and GPT4 in the game TofuKingdom}
\label{table:asym}
\end{table}

\section{Experiments}
In this section, the implementation details are presented, and we will show the game results on the three designed games.
More details can be found in Appendix and anonymous code.

\subsection{Implementation Details}
\subsubsection{Role-based Multi-turn Conversation}
\label{sec:multi-turn}
There are many different roles in the game, and the participating models need to play the assigned role and engage in multi-turn dialogue.
Both GPT-4 and ChatGPT support role-based messages as input.
Users can create messages of three different roles, including ``user,'' ``assistant,'' and ``system.''
Therefore, for models like ChatGPT and GPT-4, their own speech should be stored as an ``assistant message'' in history, and all other players' speeches, including the host's speeches, should be saved as a ``user message.''
However, the general language models, including TD003, can only support text prompts as input.
So we prepend role keywords such as ``\#\#system\#\#'' and ``\#\#answerer\#\#'' to help prompt the model to generate the speech for the assigned role.
We add the current speech to the conversation history, and the role prompt, along with the updated history, will serve as input for the next round.
Figure \ref{fig:chatfig} in Appendix \ref{appendix:a} shows the input and output examples of the game Ask-Guess to better understand the difference.

\subsubsection{Private History}
In the game, it is crucial to store the model's prior thinking processes as input for the next round.
However, these thinking processes should not be visible to other players, as some roles need to hide their identities to win the game.
Therefore, we provide distinct histories for each player. 
The private history of a player contains not only the public speeches of all players in the past but also their own reasoning process in the previous rounds, which guarantees the consistency of the model's reasoning throughout the game, ensuring the model continuously optimizes its decisions to achieve long-term goals.
Details can be found in our public code and Figure \ref{fig:cot} in Appendix \ref{appendix:d}.

\subsubsection{Role of Host}
In addition to the participating models, we also introduce a game host. The host's speech is not generated by LLMs but is produced in accordance with the game's progression and rules. For example, at the beginning of TofuKingdom, the host should announce, ``The game has started now, prince, please ask each player a question''. In SpyFall, after voting, the system tallies the player with the most votes. If player 3 receives the most votes, the host will declare, ``Player 3 received the most votes, but he is not the spy; Now the game continues.'' The host's speeches will be stored in the private history of each player, serving as prompts to remind the LLMs what to do next.

\subsubsection{Chain-of-Thought}
Chain-of-thought prompting (CoT) \cite{cot} is an effective technique for probing models' reasoning capabilities by requiring them to produce a series of intermediate reasoning steps before giving the final answer. 
In the games SpyFall and TofuKingdom, we employed CoT to encourage the model to generate its thinking process before delivering speeches and making final decisions. 
For example, during the voting stage of SpyFall, we require the model to output a string formatted as a JSON dictionary containing the keys ``thought,'' ``speak,'' and ``name.'' 
The model is expected to first output its thought in the ``thought'' field, which includes the inference of the word pair and identities of other players, as well as the general ideas for the speech.
And then model further generates the value for ``speak'' and ``name'' according to the generated thought.
We demonstrate the model's output in Appendix \ref{appendix:d}.

\subsection{Selected Foundation Models}
We chose three of the most powerful and widely used foundation models to test on the games we design:
\begin{itemize}
\item \textbf{ChatGPT}: ChatGPT is currently the most widely used large language model. It is developed by OpenAI and can excellently follow various human instructions. We use the model ``gpt-3.5-turbo'' in experiments. 
    \item \textbf{GPT-4}: GPT-4 is the next generation of GPT3, with stronger conversational and reasoning ability and support for longer context input. GPT-4 is capable of performing more complex tasks and exhibits human-like abilities in many tasks. It also supports visual input.
    \item \textbf{Text-Davinci-003}: Text-Davinci-003 is a variant model of GPT-3.5 series. Its functions are almost the same as ChatGPT, but it only supports single conversations. We abbreviate it as TD003 in this paper.
\end{itemize}
Discussion of open source LLMs is in Appendix \ref{appendix:e}

\subsection{Experiments Details}

\subsubsection{Ask-Guess}
We use labels from the Cifar-100 dataset \cite{cifar100} as the words to be guesses and run 100 games for each word.
We record the occurrence of the situation mentioned in section \ref{subsec:ask-guess} in 100 trials and compute the average for all the words as the final metrics.

\subsubsection{SpyFall}
This game requires the participation of two distinct models.
We let the three selected models battle against each other in ordered pairs.
For example, the pair $(M_a,M_b)$ means that model $a$ plays the spy and $b$ play the other players.
There are a total of six different c.
For each combination and each word pair, we run the game until the number of successful trials accumulated to 30 and report the winning rate $w_{ab}$ and the average spy living round $l_{ab}$.

\begin{figure*}[t]
\centering
\includegraphics[width=\textwidth]{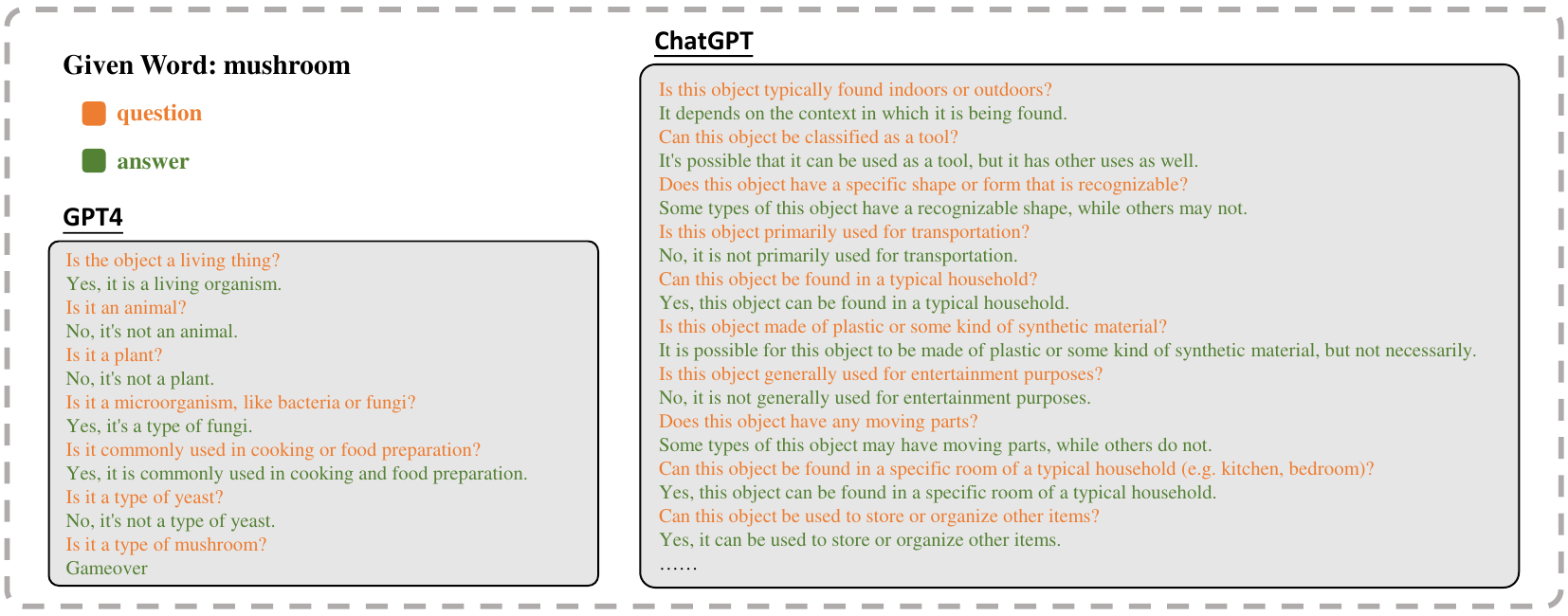}
\centering
\caption{A case to show the distinction in capabilities of ChatGPT and GPT-4 in Ask-Guess. The word to guess is mushroom.}
\label{fig:case}
\end{figure*}
\subsubsection{TofuKingdom}
There are three distinct camps and an equal number of LLMs under evaluation.
Thus, we assign one LLM to perform all roles within a single camp. 
We conduct experiments on all six possible permutations. 
For each permutation, multiple games are played until a cumulative total of 20 successful games is achieved. 
The model that plays the roles in the winning camp can get one point. 
We report the total points each model earns in all the trials.
\subsection{Quantitative Results}
\paragraph{High Discrimination Results}
In numerous prior benchmarks, the differences in capabilities between various models tend to be less distinct. For instance, the scores of ChatGPT and GPT-4 on the MMLU benchmark are 86.4 and 70. 
The overall scores on Chatbot-Arena for these two models are 1211 and 1124. 
However, the great gap between the two models when addressing complex tasks have been shown in GPT4 technical report \cite{gpt4} and some open source project like AutoGPT \cite{AutoGPT}.
Our experimental results clearly demonstrate high discrimination in the capabilities of these models.
In Ask-Guess, when providing the pre-description, the correct ratio (CR) to finish the game is 97.69, which means in the vast majority of cases, GPT-4 can guess the result within 30 rounds of Q\&A without breaking the game rules.
And GPT-4 takes only 1.57 rounds on average to correctly guess the result. 
When no pre-description is provided, ChatGPT and TD003 seem to struggle in performance.
The game can be completed successfully in just less than half of the cases.
And the average number of rounds to guess the answer is close to fifteen.
GPT-4 still holds a high 92.77 CR and only takes 40.1 rounds to guess the result.
In the game SpyFall, the gap between the models is also significant.
The winning rate of GPT-4 to TD003 is 0.70:0.13 when playing as spy and villager mutually.
GPT-4 can easily conceal its spy identity in front of TD003, but when acting as a villager, it can easily expose TD003's spy identity.
In TofuKingdom, the conversation is short, and some players may tell lies, which brings higher uncertainty to the game result, so the performance gap is relatively small. 
\paragraph{Comparisons Between ChatGPT and TD003}
As mentioned in section \ref{sec:multi-turn}, TD003 only supports single-turn text prompt input, while chatGPT has been optimized for multi-turn dialogues and designated role conversations.
So in Ask-Guess, which involves two roles combined with simple game rules, TD003 slightly outperforms ChatGPT.
But the other two games involve more roles and different camps, and the game rules are more complex, leading to a marginally lower performance compared with TD003.
Overall, TD003 can generate high-quality responses in scenarios with simple rules, but ChatGPT performs better in games with more roles and complex instructions, which is consistent with the official model description provided by OpenAI.
\subsection{Qualitative Results}

To better understand how conversational games reflect the gap in model capabilities,
we show the game dialogue in Ask-Guess without prior description in Figure \ref{fig:case}.

For each model, we create two instances with distinct role prompts to play the roles of the questioner and the answerer, respectively.
As we can see, both ChatGPT and GPT-4 can correctly understand the tasks, and they ask and answer questions according to the game rules.
However, for a given goal, GPT-4 has demonstrated an astonishing planning ability; the series of questions it asks follow a specific taxonomy.
In each round, GPT-4 shows a clear awareness of the impossible objectives that have been ruled out by previous Q\&A and ask new questions targeted at the remaining part.
However, the questions ChatGPT asks seem more disorganized and disoriented.
Although the questions it generates are helpful, it lacks a coherent taxonomy and strategic planning, so even after many rounds of conversation, ChatGPT fails to approach the correct answer.
Such high-level abilities are crucial for distinguishing the real-world intelligence of LLMs, but it is difficult to be reflected in traditional single-sample evaluations.

\section{Conclusion}

In this paper, we present GameEval, a new paradigm designed to assess the capabilities of LLMs through goal-driven conversational games. 
Unlike previous methods, GameEval needs neither annotated reference nor preference from humans or model evaluators, eliminating the need for human intervention and reducing test bias caused by evaluator models. 
GameEval requires models to apply various capabilities simultaneously in each round of conversation to reach the final goal of the assigned role, which provides a comprehensive assessment of the integrated capabilities to solve complex problems.
We designed three games of different types and evaluated three of the most representative state-of-the-art models on these games.
Experiment results show that, compared to existing methods, GameEval can better differentiate the capabilities of different LLMs.
Introducing goal-driven conversational games opens the door to evaluating the capabilities of LLMs in real-world complex scenarios.
In the near future, we will improve the game evaluation framework to accommodate smaller open-source LLM and design new games to expand GameEval.

\bibliography{aaai24}

\begin{thebibliography}{31}
\providecommand{\natexlab}[1]{#1}

\bibitem[{Aut(2023)}]{AutoGPT}
 2023.
\newblock Auto-GPT: An Autonomous GPT-4 Experiment.

\bibitem[{Chiang et~al.(2023)Chiang, Li, Lin, Sheng, Wu, Zhang, Zheng, Zhuang,
  Zhuang, Gonzalez, Stoica, and Xing}]{vicuna}
Chiang, W.-L.; Li, Z.; Lin, Z.; Sheng, Y.; Wu, Z.; Zhang, H.; Zheng, L.;
  Zhuang, S.; Zhuang, Y.; Gonzalez, J.~E.; Stoica, I.; and Xing, E.~P. 2023.
\newblock Vicuna: An Open-Source Chatbot Impressing GPT-4 with 90\%* ChatGPT
  Quality.

\bibitem[{Devlin et~al.(2018)Devlin, Chang, Lee, and Toutanova}]{bert}
Devlin, J.; Chang, M.-W.; Lee, K.; and Toutanova, K. 2018.
\newblock Bert: Pre-training of deep bidirectional transformers for language
  understanding.
\newblock \emph{arXiv preprint arXiv:1810.04805}.

\bibitem[{Dubois et~al.(2023)Dubois, Li, Taori, Zhang, Gulrajani, Ba, Guestrin,
  Liang, and Hashimoto}]{alpaca-eval}
Dubois, Y.; Li, X.; Taori, R.; Zhang, T.; Gulrajani, I.; Ba, J.; Guestrin, C.;
  Liang, P.; and Hashimoto, T.~B. 2023.
\newblock Alpacafarm: A simulation framework for methods that learn from human
  feedback.
\newblock \emph{arXiv preprint arXiv:2305.14387}.

\bibitem[{Hendrycks et~al.(2020)Hendrycks, Burns, Basart, Zou, Mazeika, Song,
  and Steinhardt}]{MMLU}
Hendrycks, D.; Burns, C.; Basart, S.; Zou, A.; Mazeika, M.; Song, D.; and
  Steinhardt, J. 2020.
\newblock Measuring massive multitask language understanding.
\newblock \emph{arXiv preprint arXiv:2009.03300}.

\bibitem[{Hendrycks et~al.(2021)Hendrycks, Burns, Kadavath, Arora, Basart,
  Tang, Song, and Steinhardt}]{math}
Hendrycks, D.; Burns, C.; Kadavath, S.; Arora, A.; Basart, S.; Tang, E.; Song,
  D.; and Steinhardt, J. 2021.
\newblock Measuring Mathematical Problem Solving With the MATH Dataset.
\newblock \emph{Cornell University - arXiv,Cornell University - arXiv}.

\bibitem[{Huang et~al.(2023)Huang, Bai, Zhu, Zhang, Zhang, Su, Liu, Lv, Zhang,
  Lei, Qi, Fu, Sun, and He}]{c-eval}
Huang, Y.; Bai, Y.; Zhu, Z.; Zhang, J.; Zhang, J.; Su, T.; Liu, J.; Lv, C.;
  Zhang, Y.; Lei, J.; Qi, F.; Fu, Y.; Sun, M.; and He, J. 2023.
\newblock C-EVAL: A Multi-Level Multi-Discipline Chinese Evaluation Suite for
  Foundation Models.

\bibitem[{Krizhevsky, Hinton et~al.(2009)}]{cifar100}
Krizhevsky, A.; Hinton, G.; et~al. 2009.
\newblock Learning multiple layers of features from tiny images.

\bibitem[{Li et~al.(2023)Li, Song, Yu, Yu, Li, Huang, and Li}]{API-Bank}
Li, M.; Song, F.; Yu, B.; Yu, H.; Li, Z.; Huang, F.; and Li, Y. 2023.
\newblock API-Bank: A Benchmark for Tool-Augmented LLMs.

\bibitem[{Liang et~al.(2022)Liang, Bommasani, Lee, and Tsipras}]{human1}
Liang, P.; Bommasani, R.; Lee, T.; and Tsipras. 2022.
\newblock Holistic Evaluation of Language Models.

\bibitem[{Lin(2004)}]{rouge}
Lin, C.-Y. 2004.
\newblock Rouge: A package for automatic evaluation of summaries.
\newblock In \emph{Text summarization branches out}, 74--81.

\bibitem[{Lin and Chen(2023)}]{llm-eval}
Lin, Y.-T.; and Chen, Y.-N. 2023.
\newblock LLM-Eval: Unified Multi-Dimensional Automatic Evaluation for
  Open-Domain Conversations with Large Language Models.
\newblock \emph{arXiv preprint arXiv:2305.13711}.

\bibitem[{Liu, Jin, and Ren(2023)}]{m3ke}
Liu, C.; Jin, R.; and Ren. 2023.
\newblock M3KE: A Massive Multi-Level Multi-Subject Knowledge Evaluation
  Benchmark for Chinese Large Language Models.

\bibitem[{Novikova et~al.(2017)Novikova, Dušek, Cercas~Curry, and
  Rieser}]{human2}
Novikova, J.; Dušek, O.; Cercas~Curry, A.; and Rieser, V. 2017.
\newblock Why We Need New Evaluation Metrics for NLG.
\newblock In \emph{Proceedings of the 2017 Conference on Empirical Methods in
  Natural Language Processing}.

\bibitem[{OpenAI(2023)}]{gpt4}
OpenAI. 2023.
\newblock GPT-4 Technical Report.
\newblock arXiv:2303.08774.

\bibitem[{Papineni et~al.(2002)Papineni, Roukos, Ward, and Zhu}]{BLEU}
Papineni, K.; Roukos, S.; Ward, T.; and Zhu, W.-J. 2002.
\newblock Bleu: a method for automatic evaluation of machine translation.
\newblock In \emph{Proceedings of the 40th annual meeting of the Association
  for Computational Linguistics}, 311--318.

\bibitem[{Pham et~al.(2023)Pham, Yang, Sheng, Zhao, Lee, Jiang, Dong, Guan, and
  Ming}]{openllm}
Pham, A.; Yang, C.; Sheng, S.; Zhao, S.; Lee, S.; Jiang, B.; Dong, F.; Guan,
  X.; and Ming, F. 2023.
\newblock {OpenLLM: Operating LLMs in production}.

\bibitem[{Qin et~al.(2023)Qin, Liang, Ye, Zhu, Yan, Lu, Lin, Cong, Tang, Qian,
  Zhao, Tian, Xie, Zhou, Gerstein, Li, Liu, and Sun}]{toolbench}
Qin, Y.; Liang, S.; Ye, Y.; Zhu, K.; Yan, L.; Lu, Y.; Lin, Y.; Cong, X.; Tang,
  X.; Qian, B.; Zhao, S.; Tian, R.; Xie, R.; Zhou, J.; Gerstein, M.; Li, D.;
  Liu, Z.; and Sun, M. 2023.
\newblock ToolLLM: Facilitating Large Language Models to Master 16000+
  Real-world APIs.
\newblock arXiv:2307.16789.

\bibitem[{Radford et~al.(2018)Radford, Narasimhan, Salimans, Sutskever
  et~al.}]{gpt}
Radford, A.; Narasimhan, K.; Salimans, T.; Sutskever, I.; et~al. 2018.
\newblock Improving language understanding by generative pre-training.

\bibitem[{Srivastava et~al.(2022)Srivastava, Rastogi, Rao, Shoeb, Abid, Fisch,
  Brown, Santoro, Gupta, Garriga-Alonso et~al.}]{bigbench}
Srivastava, A.; Rastogi, A.; Rao, A.; Shoeb, A. A.~M.; Abid, A.; Fisch, A.;
  Brown, A.~R.; Santoro, A.; Gupta, A.; Garriga-Alonso, A.; et~al. 2022.
\newblock Beyond the imitation game: Quantifying and extrapolating the
  capabilities of language models.
\newblock \emph{arXiv preprint arXiv:2206.04615}.

\bibitem[{Touvron et~al.(2023{\natexlab{a}})Touvron, Lavril, Izacard, Martinet,
  Lachaux, Lacroix, Rozi{\`e}re, Goyal, Hambro, Azhar et~al.}]{llama}
Touvron, H.; Lavril, T.; Izacard, G.; Martinet, X.; Lachaux, M.-A.; Lacroix,
  T.; Rozi{\`e}re, B.; Goyal, N.; Hambro, E.; Azhar, F.; et~al.
  2023{\natexlab{a}}.
\newblock Llama: Open and efficient foundation language models.
\newblock \emph{arXiv preprint arXiv:2302.13971}.

\bibitem[{Touvron et~al.(2023{\natexlab{b}})Touvron, Martin, Stone, Albert,
  Almahairi, Babaei, Bashlykov, Batra, Bhargava, Bhosale et~al.}]{llama2}
Touvron, H.; Martin, L.; Stone, K.; Albert, P.; Almahairi, A.; Babaei, Y.;
  Bashlykov, N.; Batra, S.; Bhargava, P.; Bhosale, S.; et~al.
  2023{\natexlab{b}}.
\newblock Llama 2: Open foundation and fine-tuned chat models.
\newblock \emph{arXiv preprint arXiv:2307.09288}.

\bibitem[{Wang et~al.(2019)Wang, Pruksachatkun, Nangia, Singh, Michael, Hill,
  Levy, and Bowman}]{SuperGLUE}
Wang, A.; Pruksachatkun, Y.; Nangia, N.; Singh, A.; Michael, J.; Hill, F.;
  Levy, O.; and Bowman, S. 2019.
\newblock SuperGLUE: A Stickier Benchmark for General-Purpose Language
  Understanding Systems.
\newblock \emph{arXiv: Computation and Language,arXiv: Computation and
  Language}.

\bibitem[{Wang et~al.(2018)Wang, Singh, Michael, Hill, Levy, and Bowman}]{GLUE}
Wang, A.; Singh, A.; Michael, J.; Hill, F.; Levy, O.; and Bowman, S.~R. 2018.
\newblock GLUE: A multi-task benchmark and analysis platform for natural
  language understanding.
\newblock \emph{arXiv preprint arXiv:1804.07461}.

\bibitem[{Wei et~al.(2022)Wei, Wang, Schuurmans, Bosma, Xia, Chi, Le, Zhou
  et~al.}]{cot}
Wei, J.; Wang, X.; Schuurmans, D.; Bosma, M.; Xia, F.; Chi, E.; Le, Q.~V.;
  Zhou, D.; et~al. 2022.
\newblock Chain-of-thought prompting elicits reasoning in large language
  models.
\newblock \emph{Advances in Neural Information Processing Systems}, 35:
  24824--24837.

\bibitem[{Zhang, Zhao, and LeCun(2015)}]{agnews}
Zhang, X.; Zhao, J.; and LeCun, Y. 2015.
\newblock Character-level convolutional networks for text classification.
\newblock \emph{Advances in neural information processing systems}, 28.

\bibitem[{Zheng et~al.(2023{\natexlab{a}})Zheng, Chiang, Sheng, Zhuang, Wu,
  Zhuang, Lin, Li, Li, Xing et~al.}]{MT-Bench}
Zheng, L.; Chiang, W.-L.; Sheng, Y.; Zhuang, S.; Wu, Z.; Zhuang, Y.; Lin, Z.;
  Li, Z.; Li, D.; Xing, E.; et~al. 2023{\natexlab{a}}.
\newblock Judging LLM-as-a-judge with MT-Bench and Chatbot Arena.
\newblock \emph{arXiv preprint arXiv:2306.05685}.

\bibitem[{Zheng et~al.(2023{\natexlab{b}})Zheng, Chiang, Sheng, Zhuang, Wu,
  Zhuang, Lin, Li, Li, Xing, Zhang, Gonzalez, and Stoica}]{chatbot}
Zheng, L.; Chiang, W.-L.; Sheng, Y.; Zhuang, S.; Wu, Z.; Zhuang, Y.; Lin, Z.;
  Li, Z.; Li, D.; Xing, E.~P.; Zhang, H.; Gonzalez, J.~E.; and Stoica, I.
  2023{\natexlab{b}}.
\newblock Judging LLM-as-a-judge with MT-Bench and Chatbot Arena.
\newblock arXiv:2306.05685.

\bibitem[{Zhong et~al.(2023)Zhong, Cui, Guo, Liang, Lu, Wang, Saied, Chen, and
  Duan}]{agieval}
Zhong, W.; Cui, R.; Guo, Y.; Liang, Y.; Lu, S.; Wang, Y.; Saied, A.; Chen, W.;
  and Duan, N. 2023.
\newblock Agieval: A human-centric benchmark for evaluating foundation models.
\newblock \emph{arXiv preprint arXiv:2304.06364}.

\bibitem[{Zhu et~al.()Zhu, Wang, Zhou, Wang, Chen, Wang, Yang, Ye, Gong, Zhang,
  Xie, Research, Textbugger, and Bertattack}]{promptbench}
Zhu, K.; Wang, J.; Zhou, J.; Wang, Z.; Chen, H.; Wang, Y.; Yang, L.; Ye, W.;
  Gong, N.; Zhang, Y.; Xie, X.; Research, M.; Textbugger, D.; and Bertattack,
  T. ????
\newblock PromptBench: Towards Evaluating the Robustness of Large Language
  Models on Adversarial Prompts.

\bibitem[{Ziems et~al.(2023)Ziems, Held, Shaikh, Chen, Zhang, and
  Yang}]{human3}
Ziems, C.; Held, W.; Shaikh, O.; Chen, J.; Zhang, Z.; and Yang, D. 2023.
\newblock Can Large Language Models Transform Computational Social Science?

\end{thebibliography}

\appendix

\clearpage

\appendix
\label{appendix}
\section{Role-based multi-turn dialogue}
\label{appendix:a}
\begin{figure*}[htb]
\centering
\subfigure[TD003]{
    \begin{minipage}{0.85\linewidth}
        \centering
        \includegraphics[width=\textwidth]{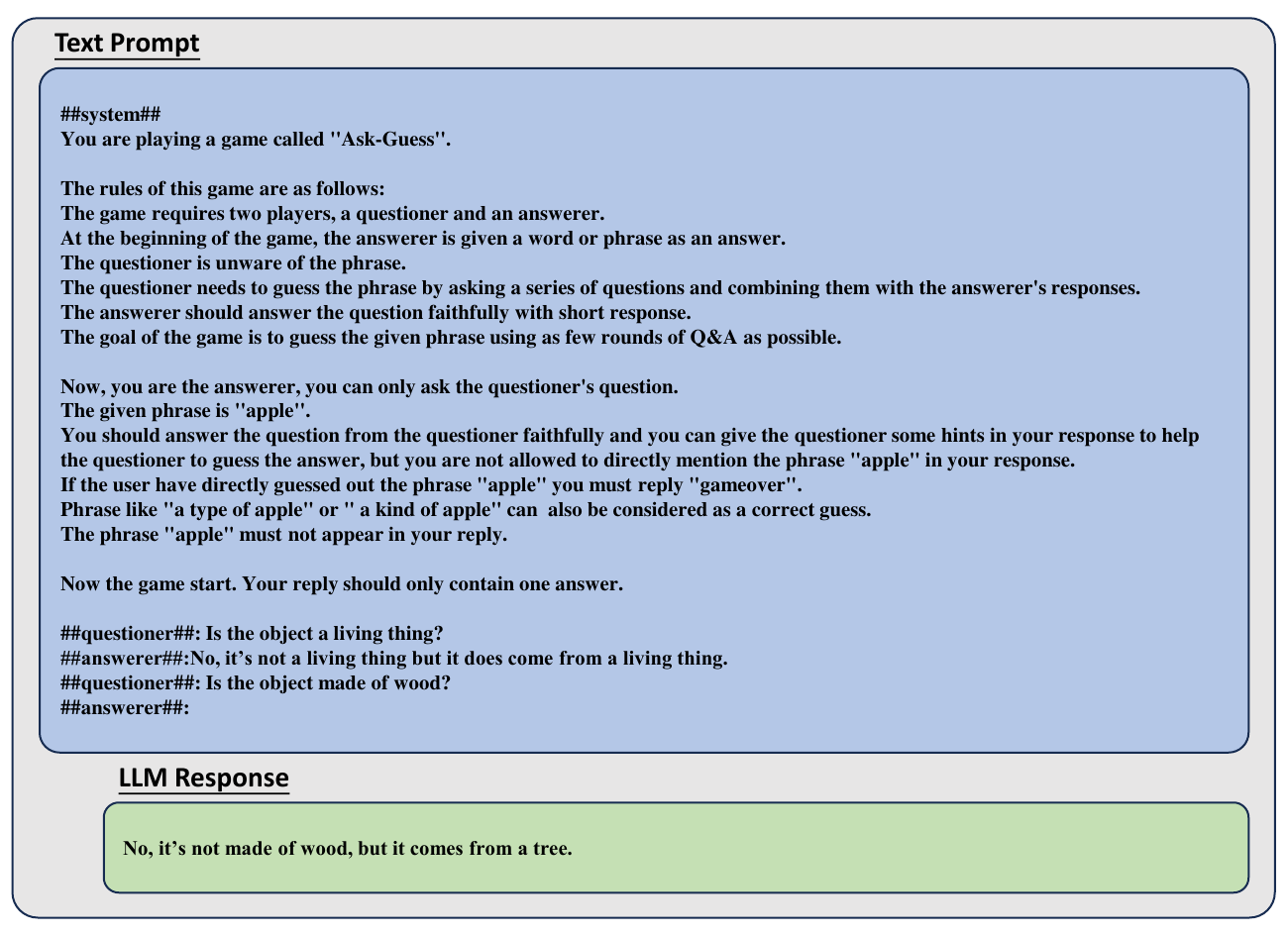}
    \end{minipage}
    \label{chatfig:a}
}
\subfigure[ChatGPT]{
    \begin{minipage}{0.85\linewidth}
        \centering
        \includegraphics[width=\textwidth]{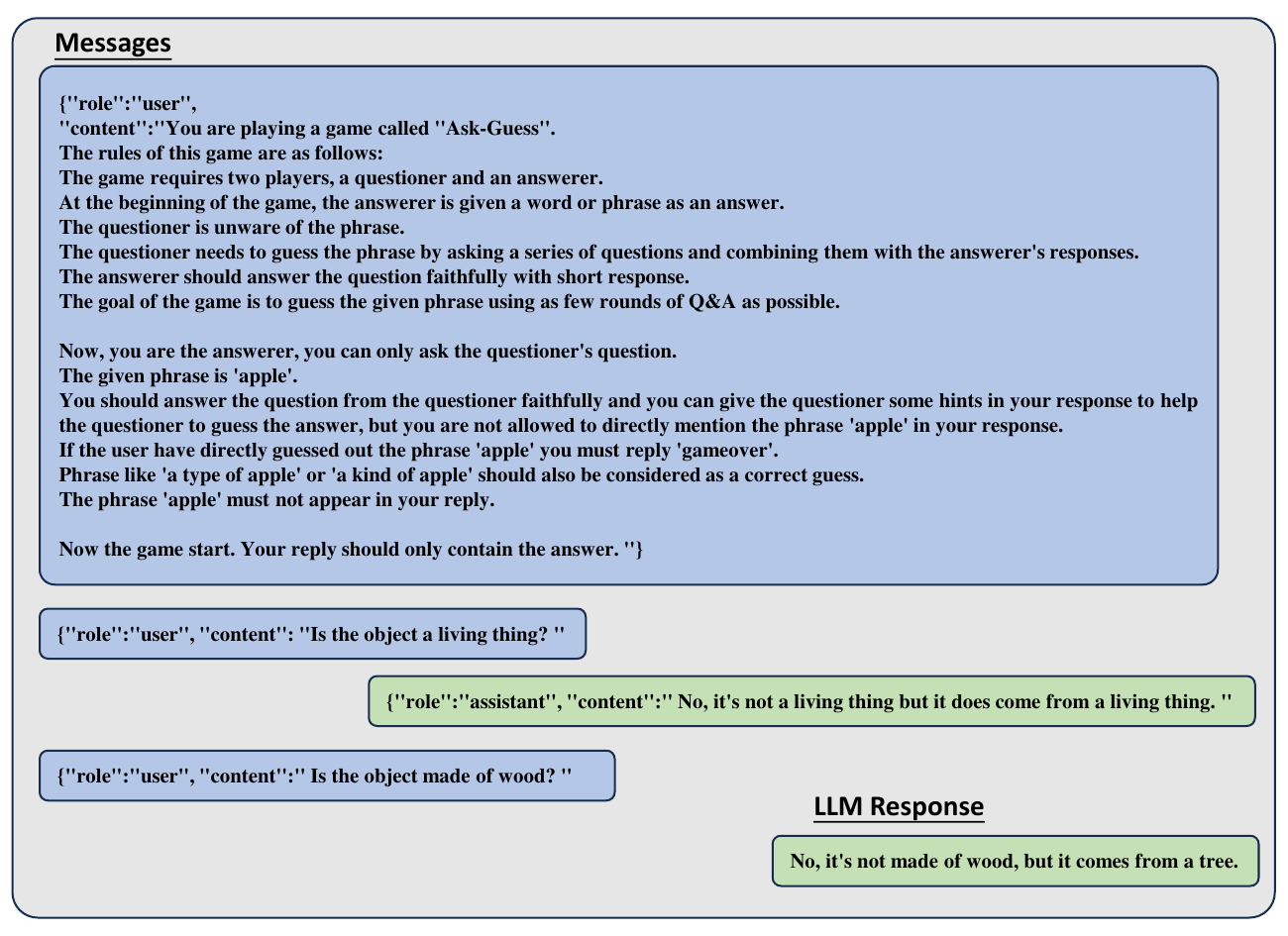}
    \end{minipage}
    \label{chatfig:b}
}

\centering
\caption{We demonstrate the different input format for different types of LLMs in game Ask-Guess. (a) Pure text prompt for the common generative LLM. (b) Role-based messages for multi-turn chat models like ChatGPT.}
\label{fig:chatfig}
\vspace{-6mm}
\end{figure*}

Common generation models typically support single-turn dialogues, accepting text input and providing text output. In contrast, some others adopt role-based messages as the input, so we have implemented two different types of game dialogue methods to simulate multi-turn conversational games.
Figure \ref{fig:chatfig} show the text input for different models in the game Ask-Guess.
Figure \ref{chatfig:a} shows the text prompt for the common generative LLM, including Text-Davinci-003 or other open-source models like alpaca or vicuna.
Figure \ref{chatfig:b} demonstrates the input for some role-based multi-turn dialogue models like ChatGPT or GPT4.



\section{Word pairs chosen in SpyFall}
\begin{figure*}[htbp!]
\centering

\subfigure[TD003 vs GPT-4]{
    \hspace{-1.5cm} 
    \begin{minipage}[t]{1.15\linewidth}
        \includegraphics[width=\textwidth]{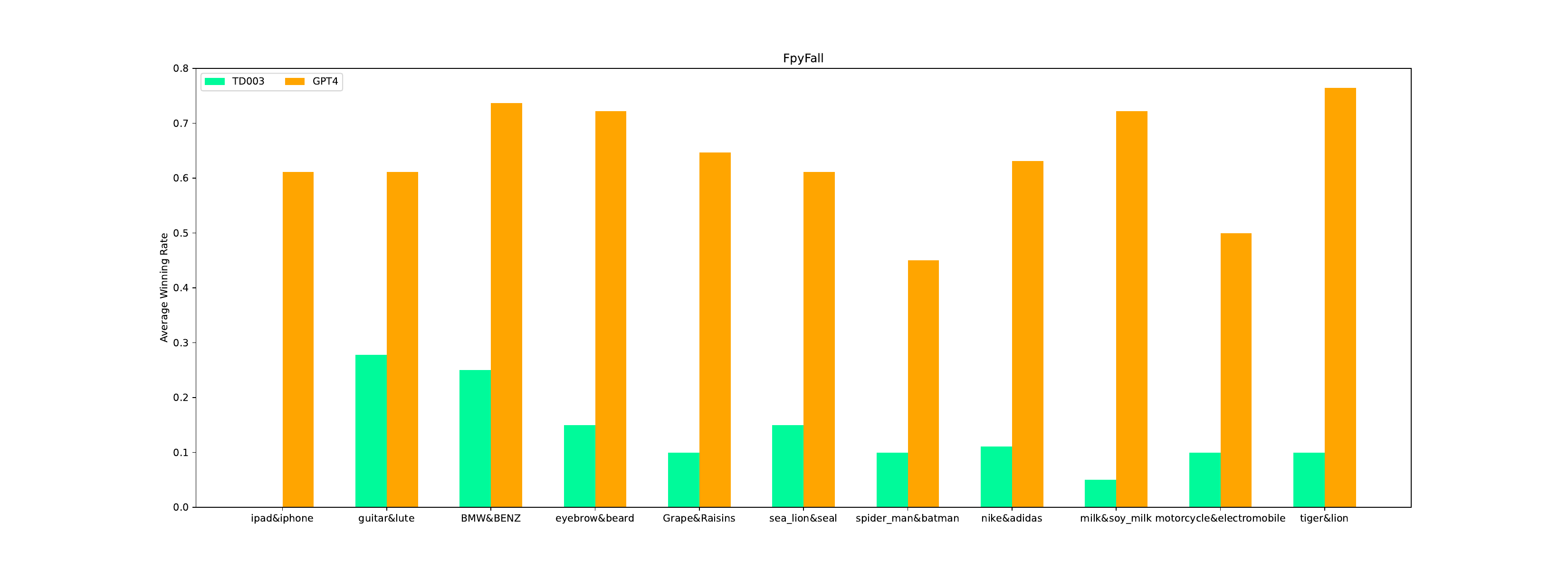}
    \end{minipage}
    \label{mainfig:1}
}
\subfigure[ChatGPT vs GPT-4]{
    \hspace{-1.5cm} 
    \begin{minipage}[t]{1.15\linewidth}
        \includegraphics[width=\textwidth]{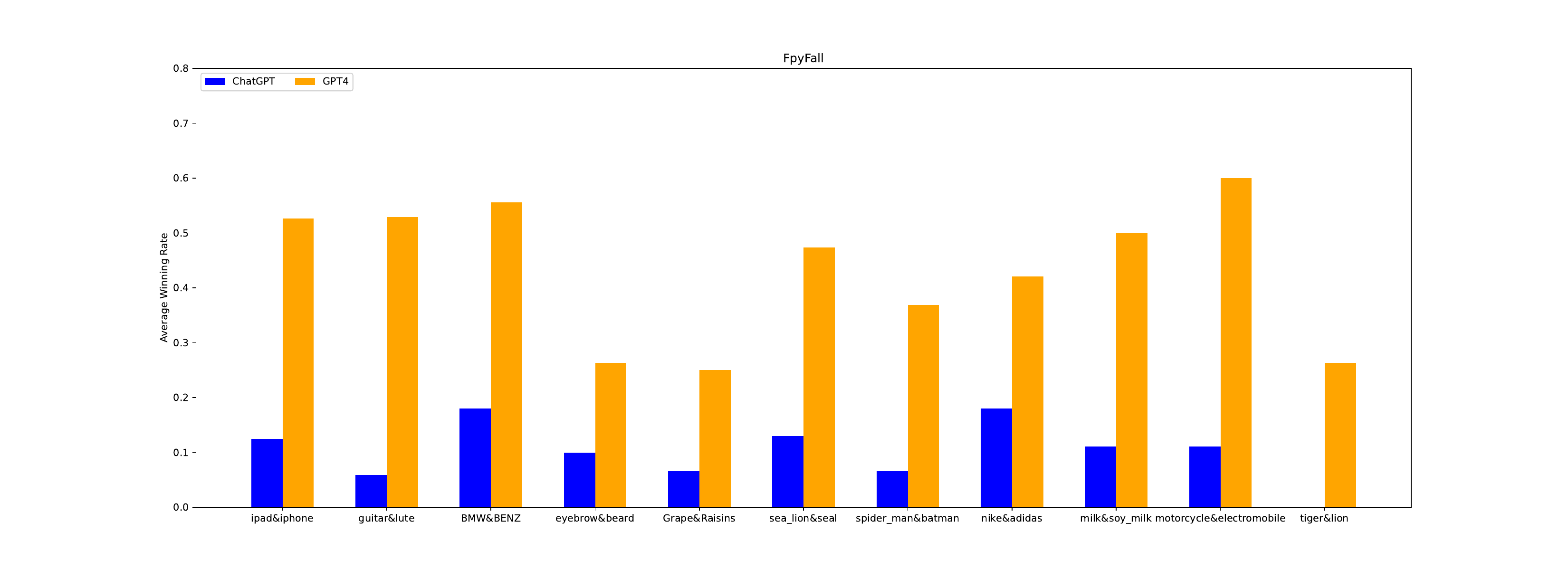}
    \end{minipage}
    \label{mainfig:1}
}

\subfigure[TD003 vs ChatGPT]{
    \hspace{-1.5cm} 
    \begin{minipage}[t]{1.15\linewidth}
        \includegraphics[width=\textwidth]{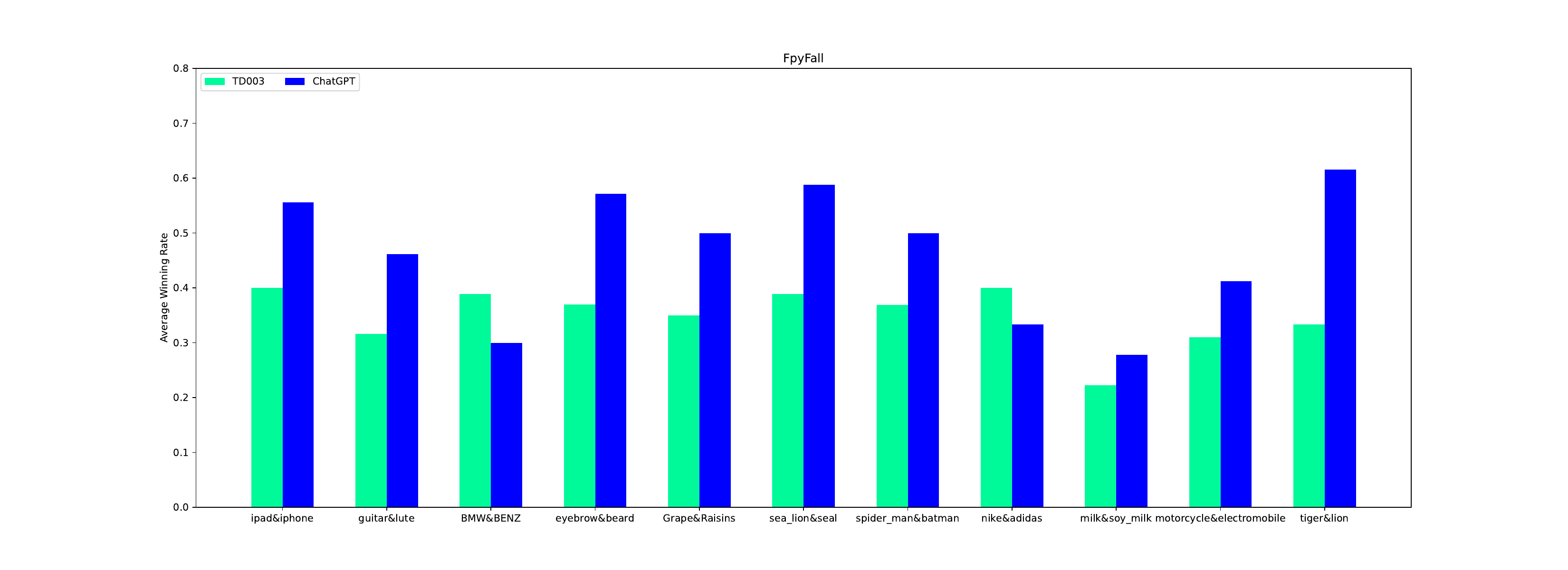}
    \end{minipage}
    \label{mainfig:1}
}
\centering
\caption{Results on the 11 word pairs in SpyFall.}
\label{fig:spywords}
\end{figure*}

For the game SpyFall, we conducted experiments on 11 pairs of related words.
Figure \ref{fig:spywords} demonstrate the game results between pair of models on each word pair.
The bar chart value represents the winning rate when playing the role of spy against the other model.
The word pairs are chosen from different domains.
For example, some word pairs can be similar electronic products like `iphone\&ipad,' and some can be different brands of similar products like `BMW\&BENZ' or `nike\&adidas.'
The models need to have a comprehensive and fine-grained understanding of these words in order to perform better in the game.
For each word pair, the first word is the spy word, and the second is the common word.
\section{Chain-of-Thought}
\label{appendix:d}
\begin{figure*}[htbp!]
\centering
        \includegraphics[width=0.75\textwidth]{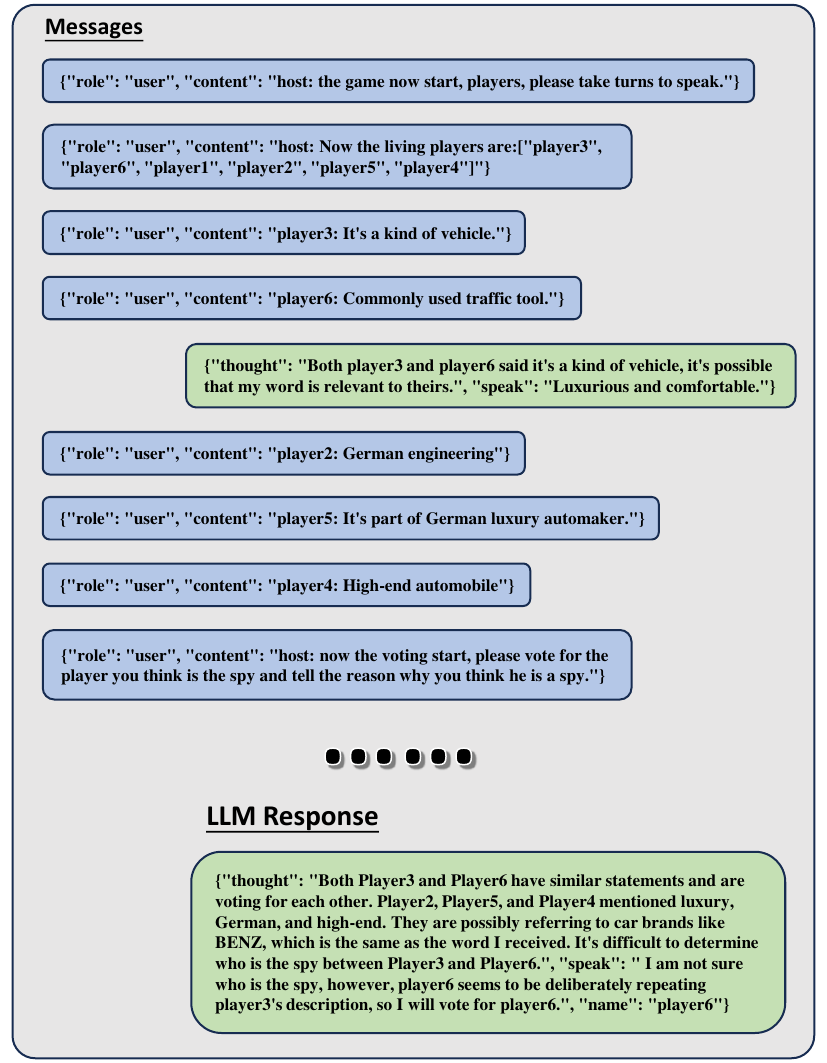}
\caption{Illustration of private history and the model's output with CoT in SpyFall.}
\label{fig:cot}
\end{figure*}

We use chain-of-thought prompting to encourage the model first analyze the previous dialogue and then make a public speech and vote according to his analysis.
Figure \ref{fig:cot} illustrate the model's output prompted by chain-of-thought in SpyFall.
In the demonstrated trial, the spy word is `BMW,' and the common word is `BENZ.' The player who is now speaking is player1, player5 is GPT4, who receives the spy word, and others are ChatGPT. 

\section{Other Details}
For TD003, we use the Completion API, which generates a response based on a text prompt. 
For ChatGPT and GPT-4, we use the Chat Completion API, which generates a response based on role-based messages.
For all the models, parameters are set to default value.
The default temperature is 1.0.
Since we run all the games multiple times, the default temperature can bring more diverse conversations.
The default maximum number of tokens for ChatGPT, TD003, and GPT4 are 4096,4097 and 32768.

\section{Limitations and Future}
\label{appendix:e}
One limitation of this paper is the absence of experimental results for open-source large language models (LLMs). 
We attempted to use the 7b-sized LLaMA2-chat \cite{llama2} and Vicuna models in the game. But our observations on the game's dialogue history revealed that these models struggled to fit themselves into their roles. They tended to forget the initial instructions and role prompt, particularly after several rounds of conversation, we have to constantly remind them who they are and what to do by inserting prompts.
For larger models, we do not have enough computation resources.
In the game Ask-Guess, we managed to achieve comparable results for some smaller LLMs by adjusting the prompt and continuously reminding the model. However, under a unified and concise prompt, these smaller models did not exhibit sufficient intelligence to participate normally in the game, highlighting the significant gap between the GPT family and open-source models.
We are currently working on optimizing the prompt and game process to enable all models to participate normally in simple games. We believe that as LLMs become more and more powerful, goal-driven conversational games may emerge as a new trend for testing the integrated capabilities of LLMs, compared to single-round instruction following or standard open-ended dialogues. We encourage researchers to evaluate the performance of other models using our public code or design new games to expand this novel evaluation paradigm.

\label{sec:reference_examples}
\end{document}